\documentclass[10pt,twocolumn,letterpaper]{article}

\usepackage{cvpr}              

\usepackage{graphicx}
\usepackage{amsmath}
\usepackage{amssymb}
\usepackage{booktabs}
\usepackage{algorithm}
\usepackage[noend]{algpseudocode}
\usepackage[table,xcdraw]{xcolor}

\usepackage[pagebackref,breaklinks,colorlinks]{hyperref}

\usepackage[capitalize]{cleveref}
\crefname{section}{Sec.}{Secs.}
\Crefname{section}{Section}{Sections}
\Crefname{table}{Table}{Tables}
\crefname{table}{Tab.}{Tabs.}

\def\cvprPaperID{38} 
\def\confName{CVPR}
\def\confYear{2022}

\begin{document}

\title{Simpler is Better: off-the-shelf Continual Learning \\through Pretrained Backbones}

\author{Francesco Pelosin\\
Ca' Foscari University\\
Venice, Italy\\
{\tt\small francesco.pelosin@unive.it}
}
\maketitle

\begin{abstract}
 In this short paper, we propose a baseline (off-the-shelf) for Continual Learning of Computer Vision problems, by leveraging the power of pretrained models. By doing so, we devise a simple approach achieving strong performance for most of the common benchmarks. Our approach is fast since requires no parameters updates and has minimal memory requirements (order of KBytes). In particular, the ``training'' phase reorders data and exploit the power of pretrained models to compute a class prototype and fill a memory bank. At inference time we  match the closest prototype through a knn-like approach, providing us the prediction. We will see how this naive solution can act as an off-the-shelf continual learning system. In order to better consolidate our results, we compare the devised pipeline with common CNN models and show the superiority of Vision Transformers, suggesting that such architectures have the ability to produce features of higher quality. Moreover, this simple pipeline, raises the same questions raised by previous works \cite{gdumb} on the effective progresses made by the CL community especially in the dataset considered and the usage of pretrained models. Code is live at \url{https://github.com/francesco-p/off-the-shelf-cl}
\end{abstract}

\begin{figure}[t]
	\begin{center}
		\centerline{\includegraphics[width=0.5\textwidth]{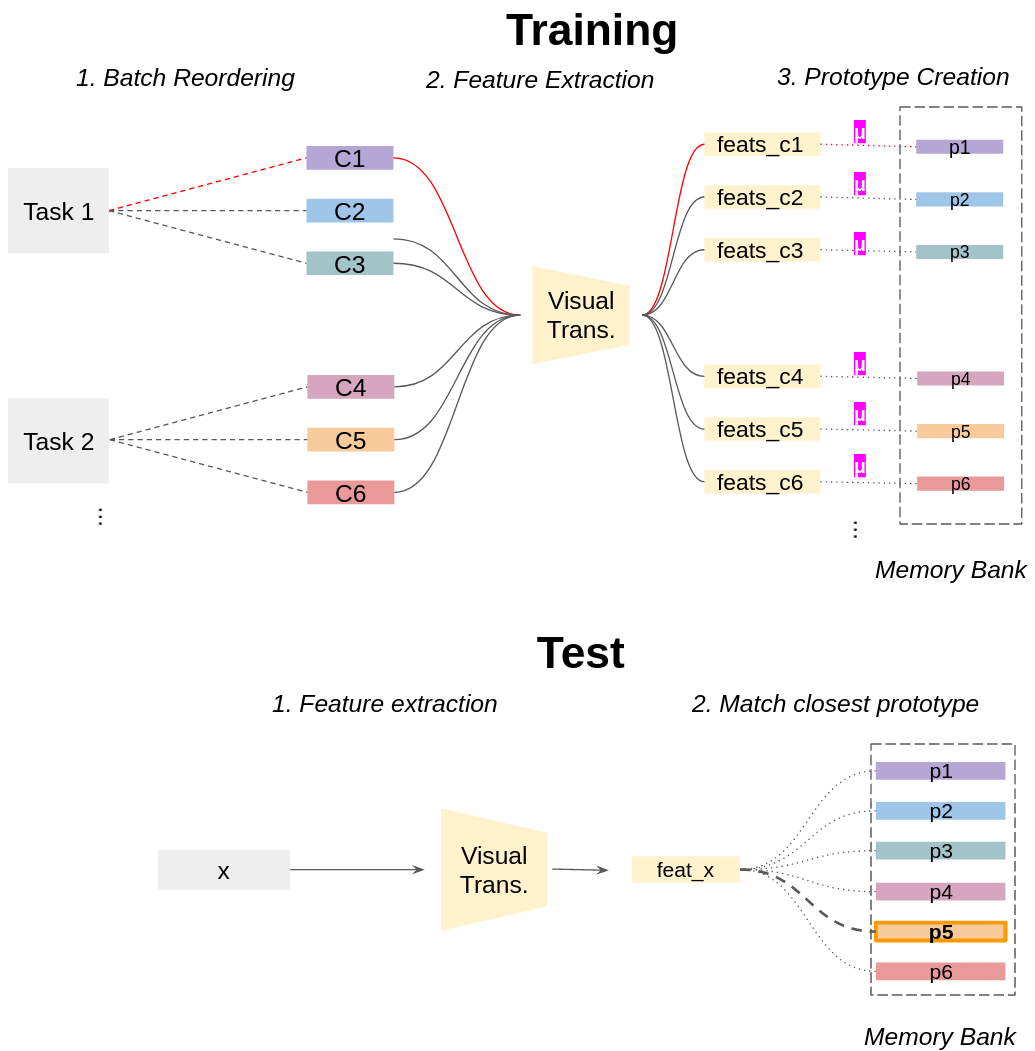}}
		\caption{Depiction of our simple baseline. Our pipeline does not perform parameters updates and consumes few KBytes as memory bank.}
		\label{fig:off-the-shelf}
	\end{center}
\end{figure}
\section{Introduction}
\label{wrk:off-the-shelf}
The need of systems able to cheaply adapt to incremental streams of data is growing. To this end, Continual Learning (CL) is receiving more and more attention by the computer vision community. With its promise of developing adaptable systems able to mimic human intelligence, the focal point of CL, lies in solving the so called \textit{catastrophic forgetting} affecting connectionists models. Catastrophic forgetting is the phenomenon where old knowledge is corrupted when the model faces incremental training sessions, this is mainly due to the distrubutional shift of the data which comes in subsequent tasks fashion. 

Until now, the CL community mainly focused in the analysis of catastrophic forgetting in Convolutional Neural Networks (CNN) models. But, as can be seen by some recent works, Vision Transformers (ViT) are asserting themselves as a valuable alternative to CNNs for computer vision tasks, sometimes, achieving better performances with respect to CNNs \cite{Chen2021WhenVT}. The power of ViTs lies in their less inductive bias \cite{DBLP:journals/corr/abs-2106-13122} and in their subsequent better generalization ability. Thanks to this ability ViTs are naturally inclined continual learners, as pointed in this recent work \cite{pelosin2022towards}.

In transformer literature, the usage of \textit{pretrained backbones} is becoming a must, in fact, training such systems requires extensive amount of data and careful hyperparameters optimization. Using pretrained backbones is common also in Computer Vision communities where CNNs are the main player. In CL literature, the pretraining is frequent, but not constant. It is typically carried on half of the analyzed dataset or through a big initial task that has the objective of facilitating the learning of low level features. The very best results, however, have been achieved when we do not skip pretraining. This can be confirmed by the CVPR 2020 Continual Learning Challenge summary report \cite{clworkshop2020}, where the authors noted that \textit{all the methods} proposed solutions leveraging pretrained backbones.

On top of that, simple baselines sometimes provide better results with respect to overly engineered CL solutions, GDumb \cite{gdumb} is such an example. In the work, the authors showed superior performance against several methods at the state-of-the-art through a system composed just by a memory random sampler and a simple learner (CNN or MLP). From a practical point of view, these methods often constitute a simple, clear, fast, intuitive and efficient solution. 

Following these lines, we explore a knn-like method to perform \textit{off-the-shelf online continual learning} leveraging the power of pretrained vision transformers. Our system constitutes a simple and memory-friendly architecture requiring zero parameters updates. Being our work one of the first using ViTs in CL, we propose a robust baseline for future works and provide an extensive comparison against CNNs.

In brevity, the contributions are the following:

\begin{itemize}
	\item We devise a \textit{simple pipeline} composed by a pretrained feature extractor and an incremental prototype bank. The latter is updated as new data is experienced. The overall cost of the method is in the storage of a pretrained backbone and few Kbytes for the memory bank.
	\item We devise a \textit{baseline for future CL methodologies} that will exploit pretrained Vision Transformers or Resnets. The baseline is fast and does not require any parameter update, yet achieving robust results in 200 lines of Python, unlocking reproducibility too.
	\item We provide a \textit{comparison for our pipeline between Resnets and Visual Transformers}. We discover that Vision Transformers produce more discriminative features, appealing also for the CL setting.
	\item In light of such results, \textit{we arise the same questions, as GDumb \cite{gdumb} does, in the progresses made by the CL community so far specifically in the quality of the datasets and in the usage of pretrained backbones.}
\end{itemize}

\begin{table*}[ht!]
	\centering
	\small
    \begin{tabular}{@{}rr|c|ccccc@{}}
\toprule
\textbf{\begin{tabular}[c]{@{}r@{}}Memory\\ KiB class\end{tabular}} & \textbf{Params} & \textbf{Model}     & \textbf{CIFAR100 \cite{cifar100}} & \textbf{CIFAR10 \cite{cifar100}} & \textbf{Core50 \cite{core50}} & \textbf{\begin{tabular}[c]{@{}c@{}}Oxford\\ Flowers102\cite{oxfordflowers}\end{tabular} } & \textbf{\begin{tabular}[c]{@{}c@{}}Tiny\\ ImgNet200\cite{imagenet}\end{tabular} } \\ \midrule
2 KiB                                                               & 11.7M           & \textit{resnet18}  & 0.53              & 0.76             & 0.72            & 0.73                                                                 & 0.55                                                              \\
2 KiB                                                               & \textit{21.8M}  & \textit{resnet34}  & 0.55              & 0.81             & 0.74            & 0.67                                                                 & 0.62                                                              \\
8 KiB                                                               & \textit{25.5M}  & \textit{resnet50}  & 0.59              & 0.80             & 0.71            & 0.70                                                                 & 0.63                                                              \\
8 KiB                                                               & \textit{60.1M}  & \textit{resnet152} & 0.67              & 0.89             & 0.72            & 0.66                                                                 & 0.76                                                              \\ \midrule
0.75 KiB                                                            & \textit{5.6M}   & \textit{ViT-T/16}  & 0.36              & 0.63             & 0.49            & 0.54                                                                 & 0.24                                                              \\
3 KiB                                                               & \textit{86.4M}  & \textit{ViT-B/16}  & 0.64              & 0.87             & 0.74            & \textbf{0.95}                                                        & 0.63                                                              \\
0.75 KiB                                                            & \textit{5.6M}   & \textit{DeiT-T/16} & 0.57              & 0.80             & 0.73            & 0.68                                                                 & 0.64                                                              \\
3 KiB                                                               & \textit{86.4M}  & \textit{DeiT-B/16} & \textbf{0.68}     & \textbf{0.90}    & \textbf{0.80}   & 0.74                                                                 & \textbf{0.79}                                                     \\ \cmidrule(lr){3-3}
\end{tabular}
	\caption{Off-the-shelf accuracy performance on different dataset benchmarks, we both analyzed a CNN model and a ViT pretrained models.}
	\label{tbl:off-the-shelf}
\end{table*}

\begin{algorithm}
\caption{Off-the-shelf CL. ``Training''}\label{alg:ots-tr}
\begin{algorithmic}
\Require $t_i, \phi, \mathcal{M}$
\For{$t_i \in \mathcal{T}$}
    \State $\mathcal{G} = \texttt{GroupByClass}(t_i)$
    \For{$g \in \mathcal{G}$}
        \State $f = \phi$(g) \Comment{Extract features}
        \State $p = \mu(f$)  \Comment{Compute mean feature}
        \State $\mathcal{M} \gets p$ \Comment{Store prototype in memory}
    \EndFor
\EndFor
\State \Return $\mathcal{M}$
\end{algorithmic}
\end{algorithm}


\section{Related Works}

Only recently few works considered self-attention models in continual learning. Li et al. \cite{li2022technical} proposed a framework for object detection exploiting Swin Transformer \cite{liu2021Swin} as pretrained backbone for a CascadeRCNN detector, the authors show that the extracted features generalize better to unseen domains hence achieving lesser forgetting rates compared
to ResNet50 \cite{resnet18} backbones. This also follows the conclusions made by Paul and Chen \cite{paul2021vision} on the fact that vision transformers are more robust learners with respect to CNNs. 

Several methods in CL use pretrained backbones as feature extractors such as in Hayes et al \cite{lda} or \cite{DBLP:conf/cvpr/AljundiKT19, 9206766} and sometimes the pretraining is carried on half (or a big portion) the dataset considered, as in PODNet \cite{podnet} or in Yu et al. \cite{Yu2021ImprovingVT}. For a more complete review on CL methodologies we point out these recent surveys \cite{review0, review1, review2}.

A similar study on pretraining for CL has been conducted by Mehta et al. \cite{Mehta2021AnEI}. In particular, they study the impact on catastrophic forgetting that a linear layer might accuse while using a pretrained backbone. Their study focuses only on Resnet18 for vision tasks, but they also include NLP tasks.

\section{Method}
\paragraph{Setting}
Continual Learning characterizes the learning by introducing the notion of subsequent tasks. In particular, the learning happens in an incremental fashion, that is, the model incrementally experiences different training sessions as time advances. Practically, a learning dataset is split in chunks where each split is considered an incremental task containing data. CL being a relatively new field, the community is still converging to a common setting notation, but we focus on an online, task-agnostic NC-type scenario. Tat is, the model forwards a pattern just once and does not have the task label at test time. As a more fine grained specific we follow \cite{core50} categorization and use a NC-type scenario where each task contains a disjoint group of classes.

More formally, given a dataset $\mathcal{D}$ and a set of $n$ disjoint tasks $\mathcal{T}$ that will be experienced sequentially:
\begin{equation}
\mathcal{T} = \left[t_1 , t_2 , \dots , t_n \right ]
\end{equation}

each task $t_i = (C_i , D_i )$ represented by a set of classes $C t = c_1^t , c_2^t \dots , c_{n^t}^t$ and training data $D_t$ (images). We assume that the classes of each task do not overlap i.e. $C^i \bigcap C^j = \emptyset$ if $i \neq j$

\paragraph{``Training'' Phase}
In the training phase, given a task $t_i \in \mathcal{T}$, a feature extractor $\phi$ and a memory bank as a dictionary $\mathcal{M}$, the procedure does the following:

\begin{enumerate}
	\item First it performs \textit{batch reordering}, that is, it groups the images of a given task by their class
	\item After grouping, it forwards each new subset to the feature extractor $\phi$
	\item Given the feature representations of a group, it computes the mean of the features to create a \textit{class prototype}
	\item Updates the memory bank $\mathcal{M}$ by storing the each computed prototype 
\end{enumerate}

At the end of the training procedure for a given task $t_i$, we would have a representative prototype vector \textit{for each class} contained in $t_i$. As we said, the prototype vector is computed as the mean feature representation of the patterns of the same class. A depiction of the ``training'' phase is reported in Figure~\ref{fig:off-the-shelf}, we also provide a pseudocode in Algorithm \ref{alg:ots-tr}. 
We also point out that there is not formal ``training'' of the network, in fact \textbf{we do not perform any parameter update}, we simply exploit the pretrained models and construct a knn-like memory system.

\paragraph{Test Phase}
After completing the training phase for a task $t_i$ the memory bank $\mathcal{M}$ will be populated by the prototypes of the classes encoundered so far. During this test phase, we simply use a \textit{knn-like approach}. Given an image $x$, the updated memory bank $\mathcal{M}$ and the feature extractor $\phi$ we devise the test phase as follows:

\begin{enumerate}
	\item Forward the test image $x$ to the feature extractor $\phi$
	\item Compute a distance between the feature representation of the image and all prototypes contained in $\mathcal{M}$
	\item We match the prototype with minimum distance and return its class
\end{enumerate}

In a nutshell, we perform k-nn with k=1 over the feature representation of an image, matching the class of the closes prototype in the bank. If the class selected is the same of the test example we would have a hit, a miss otherwise. Figure \ref{fig:off-the-shelf} reports a visual depiction of the test procedure. As distance we use a simple $l^2$, but several tests have been made with cosine similarity. Although the results with the cosine similarity are better, we opt for the $l^2$ since provides the best speedup in the implementation through Pytorch.

\section{Experiments} 
\label{sec:off-the-shelf_exp}

It is suspected that Visual Transformers generalize better with respect to CNN models. To this end, we compare CNNs models and ViTs models as feature extractors. We selected four CNN models to compare against four attention-based models. In particular, we selected DeiT-Base/15, DeiT-Tiny/15 \cite{deit}, ViT-Base/16 and ViT-Tiny/16 \cite{vit} as visual transformers. While we opted for Resnet18/34/50/152 \cite{resnet18} as CNN models. We used the \texttt{timm} \cite{timm} library to fetch the pretrained models where all the models have been trained on ImageNet \cite{imagenet} and the \texttt{continuum} \cite{continuum} library to create the incremental setting for 5 datasets, namely CIFAR10/100, Core50, OxfordFlowers102 and TinyImageNet200.

In all dataset benchmarks, we upscaled the images to $224 \times 224$ pixels in order to accommodate visual transformers which needs such imput dimension. We apply such transformation to resnet data too for a fair comparison. In order to match the closes prototype at test time, we used $l^2$ as preferred measure. 

The main results are reported in Table \ref{tbl:off-the-shelf}. The pipeline is extremely simple, yet it achieves \textit{impressive performance} as an off-the-shelf method, at cost of a very small overhead to store the prototype memory. In fact, at the end of the training phase, the memory bank translates only into \textbf{few KBytes} of storage. Although this preliminary work only consider task-agnostic setting, we remind that if at test time we are given the task label of the data, we can recast the method to work in task-aware setting. In this case, performing the test phase would be easier since the comparison of the test data will be carried only on a subset of the prototypes. On the same line, one can see that in Table \ref{tbl:off-the-shelf} we do not report each dataset task split. In fact, our method works for \textbf{any dataset split} since it just need any partition of the datasets that respect a NC protocol i.e. as long as tasks are formed by images that can be grouped in classes. We can also appreciate that transformer architectures work best in all benchmarks, suggesting direct \textit{superior generalization capabilities} with respect to CNNs or, at least, more discriminative features.

\section{Discussion} 
In light of these results, we think that this work may be extended to be considered as a baseline to assess the performance continual learning methodologies using pretrained networks as feature extractors. In particular, a thorough investigation should be carried by substituting the k-nn approach with a linear classifier, this would allow also a better comparison between resnets and visual transformers. However, we think that these preliminary results are of interest to the Vision Transformer and CL research community.

We then raise some concerns with respect to the procedure and the benchmarks used to assess new CL methodologies. As we can see, through a pretrained model, we can achieve impressive results with respect to the current CL state-of-the-art \cite{review0, review1, review2}. This point have been also raised by GDumb \cite{gdumb} where the authors questioned the progresses by providing a very simple baseline.

Moreover, we can further extend this simple pipeline to be used in \textit{unsupervised continual learning}. Actually, the extension is straightforward. In an unsupervised scenario the batch reordering step cannot be performed since we are not allowed to know each data class label. To cope with this lack of information one can substitute the step with any clustering algorithm such as K-means (we tried it but with no luck) or a more sophisticated approach such as autoencoders, self-organizing maps etc.. The test phase of the unsupervised extension would be analogous to the supervised counterpart.

\section{Conclusion}
In this short paper we proposed a baseline for continual learning methodologies that exploit pretrained  Vision Transformers and Resnets. We tackle online NC-type class-incremental learning scenario, the most common one, even though, our pipeline can be extended to different scenarios. Our off-the-shelf method is conceptually simple yet gives strong results and can be implemented in 200 lines of Python therefore enhancing reproducibility. To assess the performance of different backbones our pipeline we compared Resnets models against Vision Transformers feature extractors pretrained on the same dataset, and show that vision transformers provide more powerful features. This suggests that Vision Transformers ability to encode knowledge is is broader. Then we raise some questions about CL research progress and note that with a pretrained model and a simple pipeline one can achieve strong results and, therefore, new methodologies should drop the usage of pretrained backbones when testing on such dataset benchmarks.





{\small
\bibliographystyle{ieee_fullname}
\bibliography{egbib.bib}
}

\end{document}